\documentclass[mlmain]{jmlr}

\usepackage{caption}
\usepackage{longtable}

\usepackage{booktabs}
\usepackage[load-configurations=version-1]{siunitx} 
\usepackage{graphicx}   
\usepackage{booktabs}   
\usepackage{multirow}   
\usepackage{adjustbox}  

\theorembodyfont{\upshape}
\theoremheaderfont{\scshape}
\theorempostheader{:}
\theoremsep{\newline}

\jmlrvolume{}
\firstpageno{1}

\jmlryear{2025}
\jmlrworkshop{Symmetry and Geometry in Neural Representations}

\title[Geometric machine learning on EEG signals]{Geometric Machine Learning on EEG Signals\titlebreak}

\author{\Name{Benjamin J. Choi} \hfill
\Email{benchoi@college.harvard.edu} \\
\addr Kempner Institute at Harvard University \\
150 Western Ave, Boston, MA 02134}

\begin{document}

\maketitle

\begin{abstract}

Brain-computer interfaces (BCIs) offer transformative potential, but decoding neural signals presents significant challenges. The core premise of this paper is built around demonstrating methods to elucidate the underlying low-dimensional geometric structure present in high-dimensional brainwave data in order to assist in downstream BCI-related neural classification tasks. We demonstrate two pipelines related to electroencephalography (EEG) signal processing: (1) a preliminary pipeline removing noise from individual EEG channels, and (2) a downstream manifold learning pipeline uncovering geometric structure across networks of EEG channels. We conduct preliminary validation using two EEG datasets and situate our demonstration in the context of the BCI-relevant imagined digit decoding problem. Our preliminary pipeline uses an attention-based EEG filtration network to extract clean signal from individual EEG channels. Our primary pipeline uses a fast Fourier transform, a Laplacian eigenmap, a discrete analog of Ricci flow via Ollivier's notion of Ricci curvature, and a graph convolutional network to perform dimensionality reduction on high-dimensional multi-channel EEG data in order to enable regularizable downstream classification. Our system achieves competitive performance with existing signal processing and classification benchmarks; we demonstrate a mean test correlation coefficient of $>$0.95 at 2 dB on semi-synthetic neural denoising and a downstream EEG-based classification accuracy of 0.97 on distinguishing digit- versus non-digit thoughts. Results are preliminary and our geometric machine learning pipeline should be validated by more extensive follow-up studies; generalizing these results to larger inter-subject sample sizes, different hardware systems, and broader use cases will be crucial.

\end{abstract}

\section{Introduction}\label{sec:Introduction}

Developing systems capable of interpreting the underlying thoughts corresponding to raw brainwave patterns is an important research objective in the neural interfacing field \citep{lopez-bernal}, offering numerous applications from human rehabilitation \citep{huang} to augmentation \citep{cinel}. The use cases resulting from advanced brain-computer interfaces (BCIs) include neuroprosthetics \citep{linderman,choi3}, nonverbal communication systems \citep{moreira}, thought-based interfaces \citep{berg}, paralysis aides \citep{yoo}, and beyond.

The primary obstacle to achieving advanced neural interpretation, however, remains poor signal quality. While invasive systems (i.e., systems requiring intracranial electrode implantation) can achieve higher fidelity brainwave data, the intensive requisite open brain surgery poses a major barrier to adoption \citep{sun}. Non-invasive systems, such as electroencephalography (EEG), suffer from low signal-to-noise ratios and decreased resolution as neural signals diffuse across layers of bone and muscle tissue; electromyographic (EMG) interference remains the most prominent source of EEG contamination \citep{chen,muthukumaraswamy}. Crucially, the high dimensionality associated with multi-channel EEG systems \citep{liang} at 256 Hz and above makes robust downstream classification challenging; locating low-dimensional signal within a high-dimensional space introduces a key sparsity challenge. While some of the signal-to-noise ratio (SNR) challenges inherent in non-invasive EEG data can be remediated via hardware \citep{narasimhan}, software-based algorithmic \citep{makeig} solutions are crucial.

An underlying hypothesis guiding this paper is that we can leverage strategies from geometric machine learning (ML) to elucidate a lower-dimensional structure associated with underlying brainwave thought patterns located within high-dimensional EEG data streams—and that this geometric structure can be exploited to enable robust neural interpretation at scale. We also, however, first must handle the issue of high noise in individual EEG channels in order to unlock downstream structural exploitation. Inspired by recent studies demonstrating preliminary success with transformer-based neural denoising networks \citep{pu,yin}, we build and validate our own transformer-based network for EEG denoising. While CNN-based signal filtration algorithms take advantage of the spatial locality inherent in EEG processing \citep{schirrmeister}, transformers can also harness both the spatial locality and non-locality relevant to neural signal reconstruction \citep{cho,vaswani}. By effectively filtering EMG noise from EEG data, we hope to show preliminary evidence that transformer-based denoising can help induce geometric differentiation of EEG classes in the denoised representational space. 
Transformer-based neural denoising networks, however, do struggle with issues relating to more prolonged convergence across the parameter space (resulting in long training runs and rendering live re-training in the face of new information impractical), latency issues, and preventing unnecessary signal degradation \citep{pu,yin}. Moreover, transformer-based signal processing methods struggle from a lack of methods for effective application site determination \citep{pfeffer}. Recently, \citet{pu} achieved relative success in comparison with conventional neural network processing models but did not record a PCC above 0.9 on synthetic blind source separation (BSS) and reconstruction \citep{pu}. To iterate upon existing denoising models, we present an autoencoder-targeted adversarial transformer (AT-AT) architecture leveraging an autoencoder-based method for transformer application site determination and adversarial training for output regulation. We provide a preliminary demonstration that this AT-AT architecture can effectively support geometric differentiation of classes in the representational space on a BCI-relevant EEG classification problem. 

This paper covers a geometric machine learning-based approach to EEG signal processing; an isolated description of AT-AT for denoising is given in \citet{choi2}. Notably, we pair the upstream AT-AT signal denoising architecture with a novel downstream \emph{manifold learning} pipeline to create structure-preserving nonlinear embeddings of AT-AT-denoised EEG data. As previously alluded to, even after addressing the EEG-EMG denoising problem, one must address a curse-of-dimensionality challenge in terms of interpreting high-dimensional, multi-channel EEG data. The sparsity of high-dimensional EEG has posed major problems in terms of developing robust, regularizable EEG classification \citep{mody,hemmelmann,leon}; most BCI classification problems revolve around a low-dimensional, structured output—for example, predicting the number that a user is thinking of from EEG data. Hence, to extract low-dimensional information buried within a high-dimensional data source, one must create effective embeddings of high-dimensional EEG data \citep{wang}. However, conventional linear dimensionality reduction methods like principal component analysis (PCA) often fail to facilitate embedding space separation of class representations, as EEG signal includes nonlinearities that fall outside the scope of linear embedding strategies \citep{liang}. In this paper, we build on recent promising results in nonlinear dimensionality reduction of neural data \citep{armonaite,anuragi,pan,liang} by creating a geometric machine learning pipeline leveraging Laplacian eigenmaps, an Ollivier-Ricci curvature evolution algorithm, and graph convolutional networks (GCNs) to create EEG embeddings that exploit low-dimensional structure and enable regularizable downstream classification. For preliminary validation, we apply this pipeline on a real-world BCI use case selected because of its low-dimensional classification output structure: classifying imagined digits and non-digits from EEG data. The rationale behind the algorithmic components of our geometric ML pipeline are described in detail in \sectionref{sec:Methods2}.

\section{Methods}\label{sec:Methods}

\subsection{Building a Denoising Model}\label{sec:Methods1}

Our study uses two datasets for validation purposes: the benchmark EEGdenoiseNet \citep{zhang} dataset for validating the initial transformer-based denoising model, and the MindBigData imagined digit decoding EEG dataset \citep{vivancos} as an initial platform through which to demonstrate our geometric machine learning pipeline. In accordance with established standardized methodologies \citep{zhang} across a benchmark SNR range of -7 dB to 2 dB, we built and trained the AT-AT system (\figureref{fig:image1}) end-to-end on the EEGdenoiseNet data with an EEG-EMG BSS model objective. As alluded to in the \sectionref{sec:Introduction}, a key engineering goal of the AT-AT system was to enable faster (re-)training capabilities—essential for real-world live BCI applications. To assess training time in a live BSS environment, one must account for both the model training duration and the total time of EEG and EMG segments used. In this study, we limited our training run to just four minutes of training data in order to simulate a real-world live retraining situation; we then synthesized two training mixtures and two partitioned test mixtures at SNR levels of -7 dB and 2 dB. The model was given the subsequent objective of denoising high-variance EMG artifacts from the semi-synthetic raw EEG data.

To infer signal SNR before applying a tailored denoising model, we first employ an upstream model based on a hybrid LSTM-CNN (LC) architecture to identify the appropriate SNR target level \citep{choi} (details provided in \appendixref{apd:third}). The AT-AT system’s initial denoising pass uses a convolutional autoencoder, which has shown architectural promise in prior EEG denoising work \citep{leite}. While this autoencoder performs well in lower-noise conditions, it struggles with EEG reconstruction in higher-noise environments. To address this, we selectively invoke a time-series transformer model to handle more challenging reconstructions. However, this introduces additional complexity in the parameter space and extends training time \citep{pfeffer}, so the transformer is only activated when needed. The transformer’s invocation is determined by a heuristic based on the correlation between the autoencoder’s output and the original signal; for a sans-transformer evaluation, we refer readers to \citet{choi}. Since the autoencoder is trained to produce a noise-free EEG output, if the autoencoder output highly correlates with the raw input signal, we infer that the original signal contains less noise. Conversely, low-correlation sections ($<$0.8) are flagged for masking and time-series transformer reconstruction. While this application site determination proved fruitful in terms of correlation coefficient performance (see \sectionref{sec:Results}), a notable challenge with this hybrid approach was preserving the spectral characteristics of the EEG signal when integrating transformer outputs. Inserting transformer-generated data can cause irregularities, compromising the spectral fidelity of the model’s output. To address this, we incorporated five cycles of adversarial training; in line with prior work \citep{choi}, we found this end-to-end training pipeline enabled performance gains exceeding 10\% in terms of spectral reconstruction. The transformer-augmented autoencoder acts as the “generator,” while a 1D-CNN-based discriminator model distinguishes between real and model-generated EEG.

\subsection{Geometric ML Pipeline}\label{sec:Methods2}

Following AT-AT completion, we applied our trained denoising model to a real-world validation case: the MindBigData EEG classification problem. Previous studies have claimed promising results on inter-digit classification with this dataset \citep{mishra, mahapatra}; we attempt the task of simultaneously generalizing the imagined digit classification problem to an open-set environment. Definitions of open-set classification vary \citep{wu}, so we add clarification: in open-set learning \citep{wu}, we attempt to both classify digits while simultaneously rejecting data from non-digit, unknown classes based on two-second raw EEG samples. The open-set version of imagined digit classification has representational space class ambiguity (see \figureref{fig:image2}, left panel), and is not covered by existing MindBigData studies \citep{mishra, mahapatra}; in this paper, we attempt to leverage our AT-AT and geometric machine learning pipelines to fill a gap in current literature by demonstrating successful open-set imagined digit classification (see \sectionref{sec:Results} and \appendixref{apd:second}).  

\begin{figure}[htb]
\floatconts
  {fig:image1}
  {\caption{AT-AT processing facilitates latent space separability of neural signal classes. Pre-AT-AT t-SNE depicted left; post-AT-AT depicted right.}}
  {\includegraphics[width=1\linewidth]{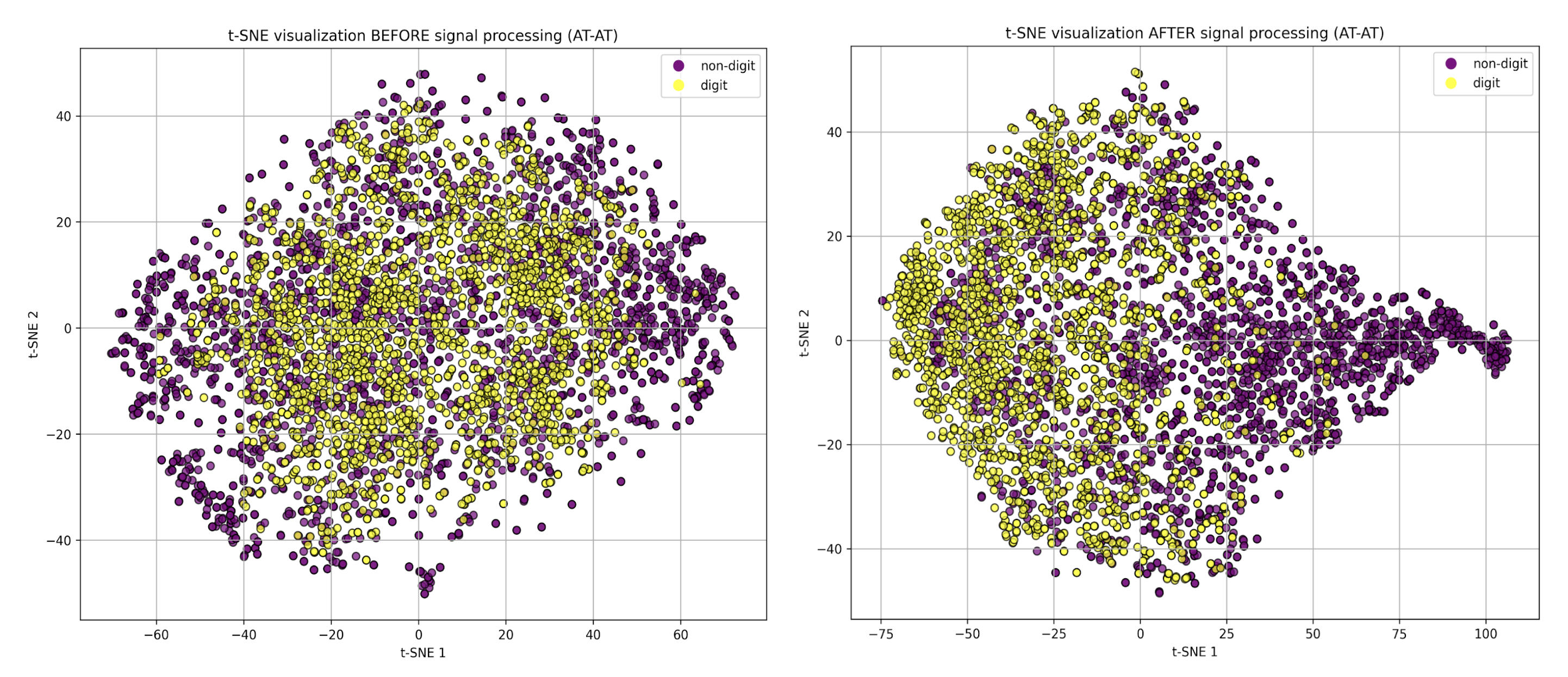}}
\end{figure}

Distinguishing between digit- and non-digit-related thoughts from raw EEG data, as in the open-set learning imagined digit classification problem, is made difficult due to lower representational space separability between these raw EEG classes (\figureref{fig:image2}, left panel). Crucially, however, after denoising the raw EEG data with AT-AT (see \sectionref{sec:Results} for more detailed metrics on AT-AT performance), AT-AT's noise reduction capabilities enabled genuine geometric representational space separability to be elucidated (\figureref{fig:image2}, right panel). The difference between the two t-distributed stochastic neighbor embedding (t-SNE) plots in \figureref{fig:image2} offer preliminary validation that beyond mere EEG-EMG source separation, our upstream AT-AT model can pull signal from noise and elucidate geometric structure in a neural interpretation context.

In order to bolster our imagined digit classification efforts while solving the aforementioned curse-of-dimensionality problem of multi-channel EEG, we implemented our primary geometric machine learning pipeline. After invoking AT-AT filtration on the individual EEG channels, we ran each channel through a one-sided FFT to create a parallel frequency vector corresponding to each time-series channel. The underlying hypothesis motivating FFT implementation was that the filtered frequency domain can elucidate key information relevant to downstream classification. After performing this FFT upscaling in conjunction with our AT-AT-denoised dataset, we next turned our focus toward tackling the second problem inhibiting successful EEG classification—that of high (multi-channel) dimensionality. 

Our first stage of dimensionality reduction involved reducing the corresponding EEG time series vectors to match the dimension of the one-sided FFT (\figureref{fig:image2}, panel 2). Given the inherent non-linearities present in EEG signal—and, crucially, the desire to preserve the local neighborhood structure \citep{liang,wang} of EEG channel nodes (as more global structures are handled via the downstream GCN)—we employed Laplacian eigenmaps \citep{belkin,liang}. The graphical representation for our spectral embedding was performed via bidirectional linking between EEG channels FP1 and FP2 along with TP9 and TP10 (corresponding to known EEG spatial patterns \citep{liu}); the subsequent reduced output at each node was paired with the corresponding one-sided FFT for further processing. Further details on our Laplacian eigenmap process are included in \appendixref{apd:first}.

\begin{figure}[htbp]
\floatconts
  {fig:image2}
  {\caption{The six manifold learning pipeline stages. (1) AT-AT filtration; (2) FFT upscaling and Laplacian eigenmap reduction; (3) graph initialization; (4) Ricci flow evolution; (5) edge cutting; (6) final representation for GCN reduction.}}
  {\includegraphics[width=0.85\linewidth]{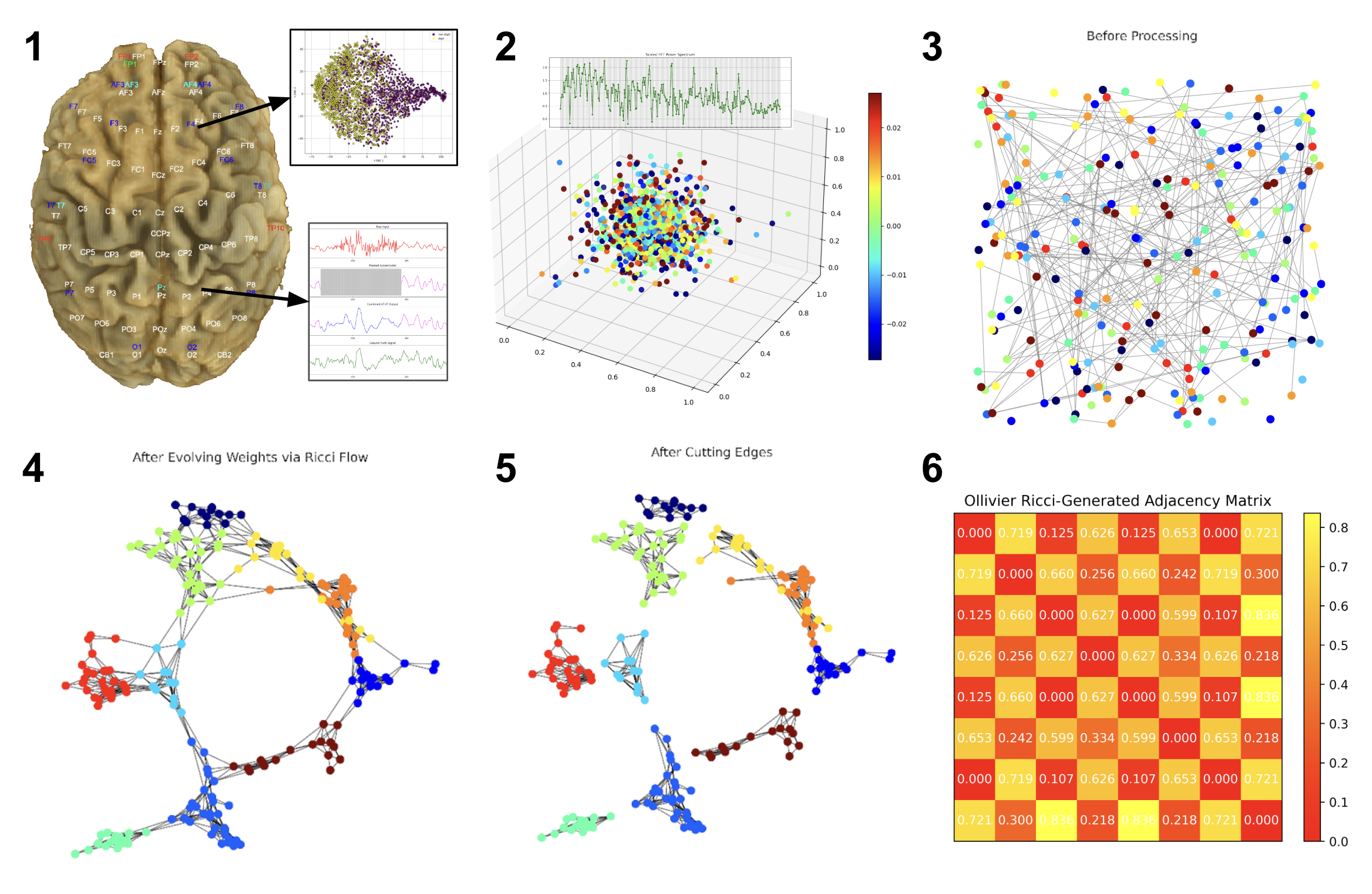}}
\end{figure}

While the combined FFT and Laplacian eigenmap representation was folded into an initial graph, we realized that manually determining edge weights and connections might not be sufficient to capture the subtle underlying community detection necessary for optimal downstream classification. Therefore, following the work of \citet{sia}, \citet{weber}, and \citet{tian}, we applied Ricci flow via Ollivier-Ricci curvature to discover more nuanced spatial relationships between EEG channels across different brain regions in both time and frequency domains. Ricci flow, a mathematical concept originating from Riemannian geometry, iteratively deforms the metric of a manifold by adjusting its curvature over time. In our context, a discrete Ricci flow analog was applied to the graph edges by leveraging Ollivier's notion of Ricci curvature \citep{ollivier}, which defines how two nearby distributions of mass (neighbors of EEG nodes) compare in terms of transport cost (using Wasserstein distance). This comparison allowed us to dynamically evolve the relationships between EEG channels as we refined the graph structure.

To effectively compare the one-sided FFT channels and Laplacian eigenmap-reduced time-series channels, we created a custom distance metric (see \appendixref{apd:first}). This metric was essential for determining relationships across the graph, given that standard Euclidean distances could not sufficiently capture the complex interdependencies between both spatial and frequency representations. The metric incorporated both local neighborhood structures and broader spectral features by combining spatial proximity (from the Laplacian eigenmap) with frequency characteristics (from the FFT), enabling a more holistic comparison between EEG channels. Starting from an initial configuration of 64,000 adjacency relationships and 1,000 hypothesized subgraphs, the Ricci flow was iteratively applied over 10 unsupervised iterations per subgraph. The evolution of edge weights under the Ricci flow was driven by the computed Ollivier-Ricci curvature for each edge, which indicated whether the relationship between two nodes (channels) was converging or diverging over time. Edges with low Ricci curvature-based  weighting indicated strong, cohesive communities (or clusters) between nodes, whereas highly-weighted edges pointed to more distant relationships. This distinction allowed us to dynamically adapt the graph structure to better reflect underlying community structures in the data.

During each iteration, edge weights were updated according to the formula:
\[
w_{uv}(t+1) = w_{uv}(t) \cdot e^{-\alpha \kappa(u,v)}
\]
where $\kappa(u,v)$ is the Ollivier-Ricci curvature of edge $(u,v)$ and $\alpha$ is the learning rate controlling how aggressively the graph’s structure evolves (see \appendixref{apd:first}). As the iterations progressed, we observed a bimodal edge weight distribution, which naturally occurred due to the distinct nature of inter-frequency, inter-time-series, and intra-channel relationships within the EEG data. The bimodality in this distribution is reflective of the temporal and spectral communities present in the graph structure, with clusters of nodes elucidated during evolution \citep{weber, sia}. After evolving the graph's edge weights over the 10 iterations, we performed edge cutting based on the edge weight distribution. The optimal 0.6 cut ratio was selected in light of the weight distribution (see \figureref{fig:image4} in \appendixref{apd:first}) to elucidate useful relationships, cutting past less informative connections while preserving subtle trends helpful for capturing meaningful community structures. This cut probability allowed us to filter out noisy edges while maintaining helpful interdependencies, resulting in a refined representation of the underlying EEG channel relationships (see \appendixref{apd:first} for further details).

\begin{figure}[htb]
\floatconts
  {fig:image3}
  {\caption{Class embeddings produced by the GCN after the manifold learning pipeline.}}
  {\includegraphics[width=0.85\linewidth]{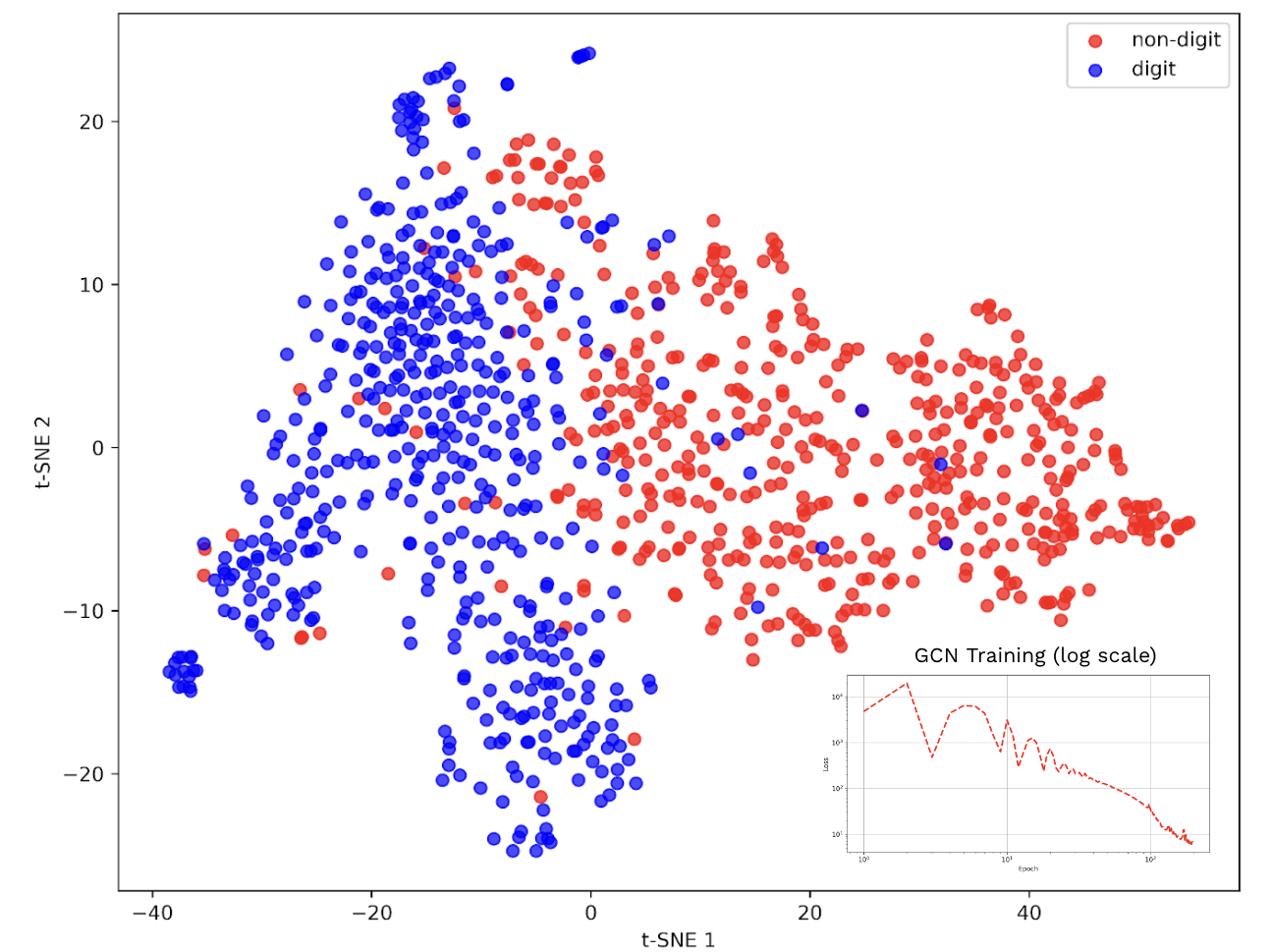}}
\end{figure}

The final composite adjacency matrix, produced by aggregating the surviving edges across the subgraphs, served as the foundation for downstream training of a graph convolutional network (GCN). This matrix, derived via the Ricci flow process, provides a nuanced representation of the EEG signal's spatial and spectral structures. We then trained a graph convolutional network (GCN) to reduce the Ricci flow-generated graph into a final embedding for downstream classification. The GCN was trained over 100 epochs to preserve distances while performing eight-to-one dimensionality reduction. Skip connections were implemented to prevent over-smoothing—as shown in the GCN training sub-panel (\figureref{fig:image3}, lower right) and the clear digit versus non-digit separability, our GCN (i.e., the culmination of our manifold learning pipeline) was able to foster clear geometric differentiation of digit versus non-digit classes (\figureref{fig:image3}). Indeed, post-manifold learning, downstream classification was relatively trivial—a lightweight 1D CNN (\appendixref{apd:second}) proved sufficient in terms of handling final classification (see \sectionref{sec:Results}). Further details on the adjacency matrix and GCN pipeline are provided in \appendixref{apd:first}.

\section{Results}\label{sec:Results}

As mentioned, upstream EEG-EMG denoising helped foster some class separability in the representational space. This was a second-order effect; AT-AT was trained solely in the context of the aforementioned EEGdenoiseNet semi-synthetic contamination problem. AT-AT posted a mean reconstructive correlation coefficient (CC) with ground truth of 0.951 at 2 dB (95\% CI: 0.947, 0.954), a temporal relative root mean square error (tRRMSE) of 0.317 (95\% CI: 0.305, 0.329), and a spectral relative root mean square error (sRRMSE) of 0.270 (95\% CI: 0.238, 0.303) on EEGdenoiseNet-based test cases. On the low end of the SNR spectrum (-7 dB), AT-AT posted a CC with ground truth of 0.703 (95\% CI: 0.679-0.726), tRRMSE of 0.759 (95\% CI: 0.731, 0.786), and sRRMSE of 0.800 (95\% CI: 0.753, 0.848). Total mean training time on a T4 High-RAM GPU was measured at 249.1 seconds.

\begin{table}[hbt]
\floatconts
  {tab:tab1}
  {\caption{AT-AT processing performance compared with existing models. A, B, and C denote \citet{zhang1}, \citet{yin}, and \citet{cui}, respectively.}}
  {\begin{tabular}{lcccc}
  \toprule
  \bfseries Model & \bfseries C-T-S (-7 dB) & \bfseries C-T-S (2 dB) & \bfseries Est. Parameters \\
  \midrule
  Novel CNN (A) & 0.69-0.72-0.65 & 0.92-0.33-0.30 & 58.7M \\
  GCTNet (B) & \textit{did not test} & 0.94-0.28-\textit{unk.} & $\sim$10M \\
    EEGIFNet (C) & \textit{did not test} & 0.95-0.32-\textit{unk.} & 5.9M \\
  AT-AT (Ours) & 0.70-0.76-0.80 & 0.95-0.32-0.27 & 438K+ \\
  \bottomrule
  \end{tabular}}
\end{table}

Our geometric machine learning pipeline achieved a 1D-CNN-based 97.0\% test accuracy (95\% CI: 93.66, 100.0) on the open-set digit- versus non-digit thought classification from two-second EEG samples. Further theoretical estimates of the performance of our geometric machine learning pipeline when combined with reported state-of-the-art inter-digit classification models are included in \appendixref{apd:second}. 

\begin{table}[hbt]
\centering
\caption{Downstream classification performance overview after manifold learning. More granular details included in \appendixref{apd:second}.}
\captionsetup{skip=0.1cm}
\label{tab:tab2}
\begin{tabular}{lc|c}
\toprule
\bfseries Metric & \bfseries Score & \bfseries Confusion Matrix \\
\midrule
Accuracy    & 0.970 & \multirow{5}{*}{\adjustbox{valign=m}{\hspace{0.4cm}\includegraphics[width=0.4\textwidth]{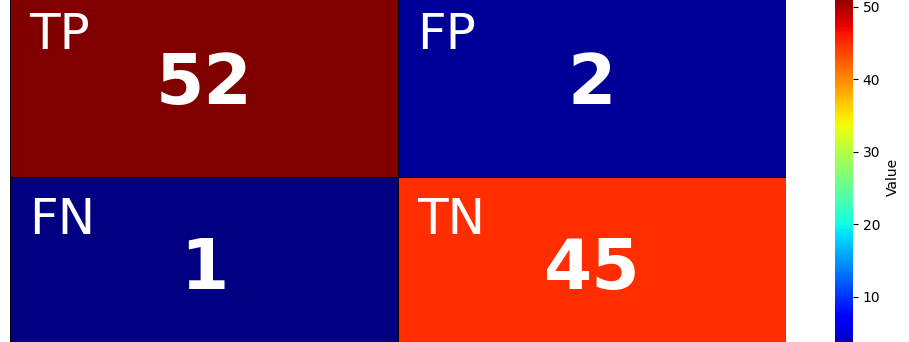}}} \\
F1 Score    & 0.968 &  \\
AUCROC      & 0.971 &  \\
Sensitivity & 0.978 &  \\
Specificity & 0.963 &  \\
\bottomrule
\end{tabular}
\end{table}

\section{Discussion}\label{sec:Discussion}

\subsection{Implications}\label{sec:Discussion1}

We can observe evidence for the efficacy of our manifold learning pipeline in the induced latent space class separability in \figureref{fig:image3}, especially when contrasted with the initial lack of embedding class differentiation in \figureref{fig:image1} (left panel). We can also see preliminary evidence for the Ollivier-Ricci curvature-based algorithm in the GCN embedding performance and rapid observed GCN convergence (\figureref{fig:image3}, lower right panel), potentially serving as an indication that evolving edge weights using Ricci flow transforms the EEG graph to elucidate intrinsic geometric relationships between node collections. By uncovering subtle underlying graph structure, this revelation of nuanced communities may facilitate the downstream GCN's ability to learn and preserve distances in the embedding space. Our preliminary validation of the upstream AT-AT denoising model also proved promising; by abstracting out high-noise segment handling and implementing selective transformer invocation, we were able to train a reasonably competitive (\tableref{tab:tab1}) model performing filtration of high-variance EMG artifacts with a relatively low parameter count.

Beyond direct neural interfacing applications in the realm of imagined digit classification (see \appendixref{apd:second} for details), the manifold learning pipeline presented in this study could open doors to further research on the underlying semantic geometric structure present in high-dimensional brainwave data. The t-SNE visualizations in \figureref{fig:image1} and \figureref{fig:image3} offer a preliminary indication that raw high-dimensional EEG signals could contain an underlying geometric structure corresponding to semantic thoughts. More extensive validation of this early idea could open the door to future studies on EEG-based thought modeling with implications for uncovering fundamental properties of intelligence.

\subsection{Limitations}\label{sec:Discussion2}

It is crucial to note that more extensive studies will be needed to generalize these results to larger sample sizes, broader user studies, different hardware systems, and wider-ranging use cases. In particular, the geometric manifold learning pipeline will require validation on a wider slate of EEG-based classification tasks; the limited deployment of our methods to imagined digit classification (see \appendixref{apd:second}) curtails the pipeline's broader applicability without further adaptation. The stability and robustness of the Ricci flow-based graph optimization also require further validation, particularly across different signal modalities or downstream use cases. Our AT-AT denoising system must also be validated across a wider slate of both semi-synthetic and real-world test settings beyond the high-variance EEGdenoiseNet EMG filtration context and subsequent imagined digit classification deployment. Importantly, while intensive and beyond the scope of this initial study (which relied on available open-source EEG data), validation of these methods will also require larger-scale recruitment of human participants for exhaustive, use-case-specific EEG data collection.

It is also important to note that the capabilities unveiled in this study—particularly the ability to potentially discern internal thoughts from non-invasive external brainwave activity—could pose a privacy risk to individuals if misused. Careful consideration, alongside the development of robust ethical and regulatory frameworks, will be vital to ensure that BCI advancements are managed responsibly and in alignment with human values.

\section*{Acknowledgements}

This work was supported by Prof. Demba Ba and the Kempner Institute for the Study of Natural and Artificial Intelligence at Harvard University, whose contributions and funding are sincerely appreciated.

\bibliography{pmlr-sample}

\appendix

\section{Manifold Learning Details}\label{apd:first}

\subsection{Laplacian Eigenmap}

For our initial dimensionality reduction, we apply a Laplacian eigenmap on networks of AT-AT-denoised channels. The adjacency matrix $\mathbf{A}$ captures pairwise relationships between rows based on specific regions of interest. Links are created between temporal-parietal (TP) cortical regions and frontal-parietal (FP) cortical regions to elucidate spatial relationships.

The graph Laplacian $\mathbf{L}$ is computed as:
\[
\mathbf{L} = \mathbf{D} - \mathbf{A}
\]
where $\mathbf{D}$ is the degree matrix, defined as:
\[
D_{ii} = \sum_{j=1}^{n} A_{ij}
\]
The spectral embedding is performed by solving the eigenvalue problem:
\[
\mathbf{L} \mathbf{v} = \lambda \mathbf{v}
\]
where $\lambda$ are the eigenvalues and $\mathbf{v}$ are the corresponding eigenvectors. The spectral embedding output is subsequently interleaved with the result of the one-sided FFT and passed to the Ricci flow stage of the manifold learning pipeline.

\subsection{Ricci Flow}

\algorithmref{{alg:ricci}} describes the mechanics of our Ricci flow algorithm for optimal graph derivation. Our custom metric capturing spatial and frequency relationships is described in \algorithmref{{alg:ricci}}, \textbf{1(b)}. \figureref{fig:image4} depicts summary distributions and statistics for our Ricci flow-driven algorithm on the imagined digit classification deployment case, with the bimodal nature of edge weight distribution (as mentioned in \sectionref{sec:Methods}) evident. Edge cut frequencies and weight aggregations are displayed in the central panels of \figureref{fig:image4}. 

\begin{algorithm}[htb]
\floatconts
{alg:ricci}
{\caption{Overview of Discrete Ricci Flow Algorithm for Neural Graph Evolution}}
{
\begin{enumerate}
  \item \textbf{Graph Initialization}
    \begin{enumerate}
      \item Define the graph by computing the initial distance between node pairs:
        \[
        d(u, v) = 
        \begin{cases} 
        \frac{1}{2}(\|f(u) - f(v)\| + \text{FFT}\Delta(f(u),f(u))) \\
        \|f(u) - f(v)\| \\
        \frac{1 - \rho(\text{FFT}(f(u)), \widehat{f(v)})}{2}
        \end{cases}
        \]
    where FFT-derived comparisons are performed using one-sided power spectra:  
    \[
    \left| \sum_{n=0}^{N-1} x_n e^{-i 2 \pi k n / N} \right|^2, \quad k = 0, 1, \dots, \frac{N}{2} - 1.
    \]  
    \end{enumerate}

  \item \textbf{Ricci Flow Evolution}
    \begin{enumerate}
      \item Compute the Ollivier-Ricci curvature for each edge:
        \[
        \kappa(u, v) = 1 - \frac{W_1(\mu_u, \mu_v)}{d(u,v)},
        \]
        where $W_1$ is the Wasserstein distance and $\mu_u, \mu_v$ are distributions over neighbors of $u$ and $v$.
      \item Evolve edge weights over time based on Ricci curvature until convergence:
        \[
        w_{uv}(t+1) = w_{uv}(t) \cdot e^{-\alpha \kappa(u,v)}, \quad \alpha > 0.
        \]
    \end{enumerate}

  \item \textbf{Post-Processing}
    \begin{enumerate}
      \item Identify and remove the top $\rho\%$ of edges by weight:
        \[
        E_{\text{cut}} = \{ (u,v) \in E : w_{uv}(t) \text{ in top } \rho\% \text{ of } E \}.
        \]
      \item Compute the aggregated adjacency matrix over all iterations:
        \[
        \bar{A} = \frac{1}{N} \sum_{i=1}^N A_i, \quad A_i \text{ is the adjacency matrix of } G_i.
        \]
    \end{enumerate}
\end{enumerate}
}
\end{algorithm}

\begin{figure}[tb]
\floatconts
  {fig:image4}
  {\caption{Summary of Ollivier-Ricci curvature pipeline metrics as described in \sectionref{sec:Methods2}.}}
  {\includegraphics[width=0.83\linewidth]{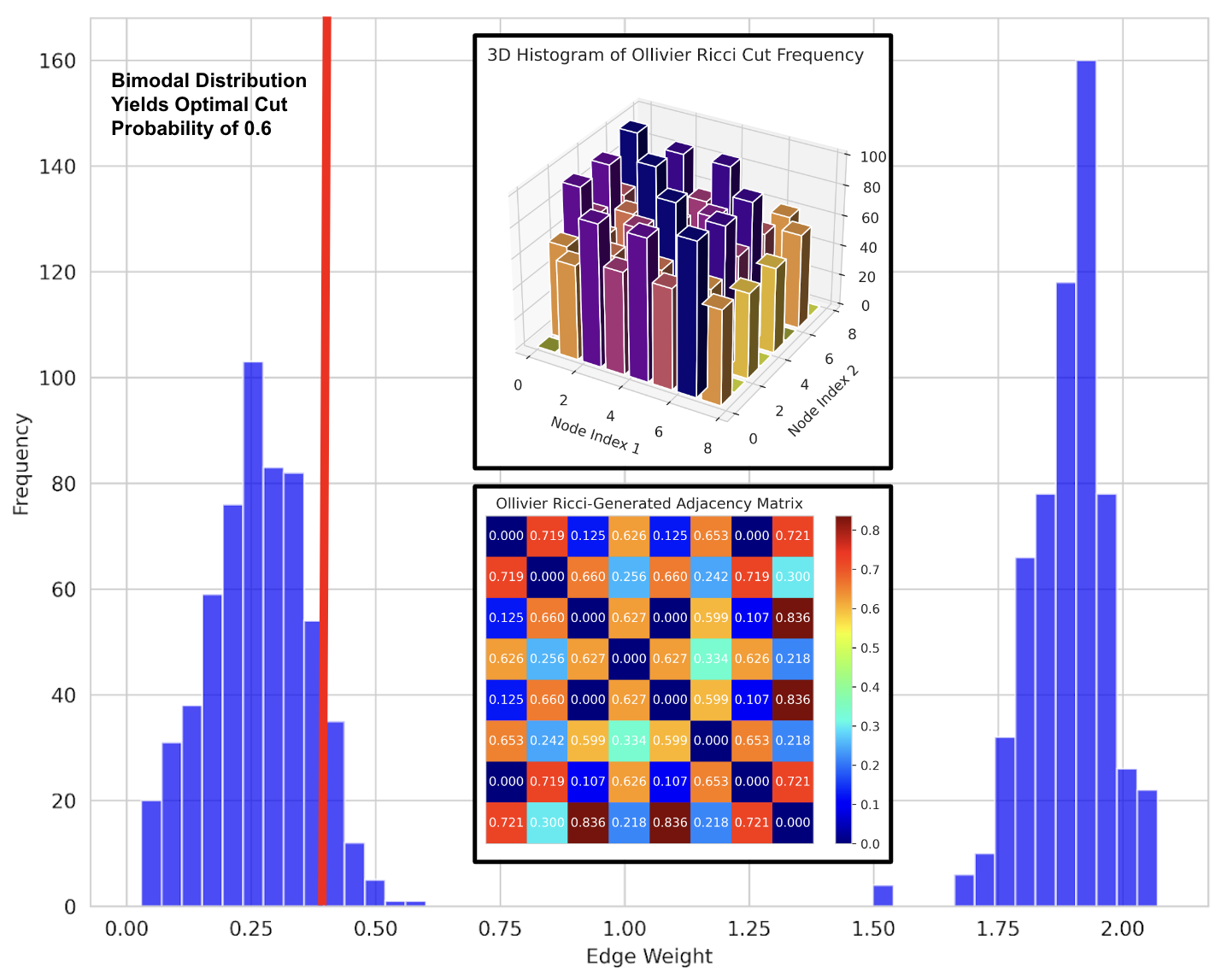}}
\end{figure}

\subsection{Graph Convolutional Network}

Architecturally, the GCN consists of two graph convolutional layers; the relatively shallow architecture helps mitigate \emph{oversmoothing}, a phenomenon where the node representations can become indistinguishable from each other after extended iterative computation \citep{peng}. The optimized adjacency matrix \( \mathbf{A} \) is derived via the Ricci flow-based algorithm, which preserves the geometric structure of the graph while enhancing the meaningful relationships between the nodes. The model is optimized with Adam with a learning rate of $1 \times 10^{-3}$. To preserve geometric relationships, the GCN is trained to maintain original distances via the objective function in the embedding space. We also implement skip connections at each layer to prevent oversmoothing, in accordance with \citet{huang2}.

The first graph convolution layer transforms the input features \( \mathbf{X} \) as:

\[
\mathbf{H}^{(1)} = \sigma\left( \mathbf{A} \mathbf{X} \mathbf{W}^{(1)} \right) + \mathbf{X} \mathbf{W}_{\text{skip}}^{(1)}
\]

The second graph convolution layer reduces the dimensionality of each node network to a vector embedding in \( \mathbb{R}^{256} \) for final 1D-CNN downstream classification:

\[
\mathbf{H}^{(2)} = \sigma\left( \mathbf{A} \mathbf{H}^{(1)} \mathbf{W}^{(2)} \right) + \mathbf{H}^{(1)} \mathbf{W}_{\text{skip}}^{(2)}
\]

\section{Further Results}\label{apd:second}

\begin{figure}[tb]
\floatconts
  {fig:image5}
  {\caption{Estimated true performance on the open-set imagined digit recognition problem with our geometric machine learning pipeline when combined with the top claimed inter-digit classification performance in literature \citep{mahapatra}.}}
  {\includegraphics[width=0.74\linewidth]{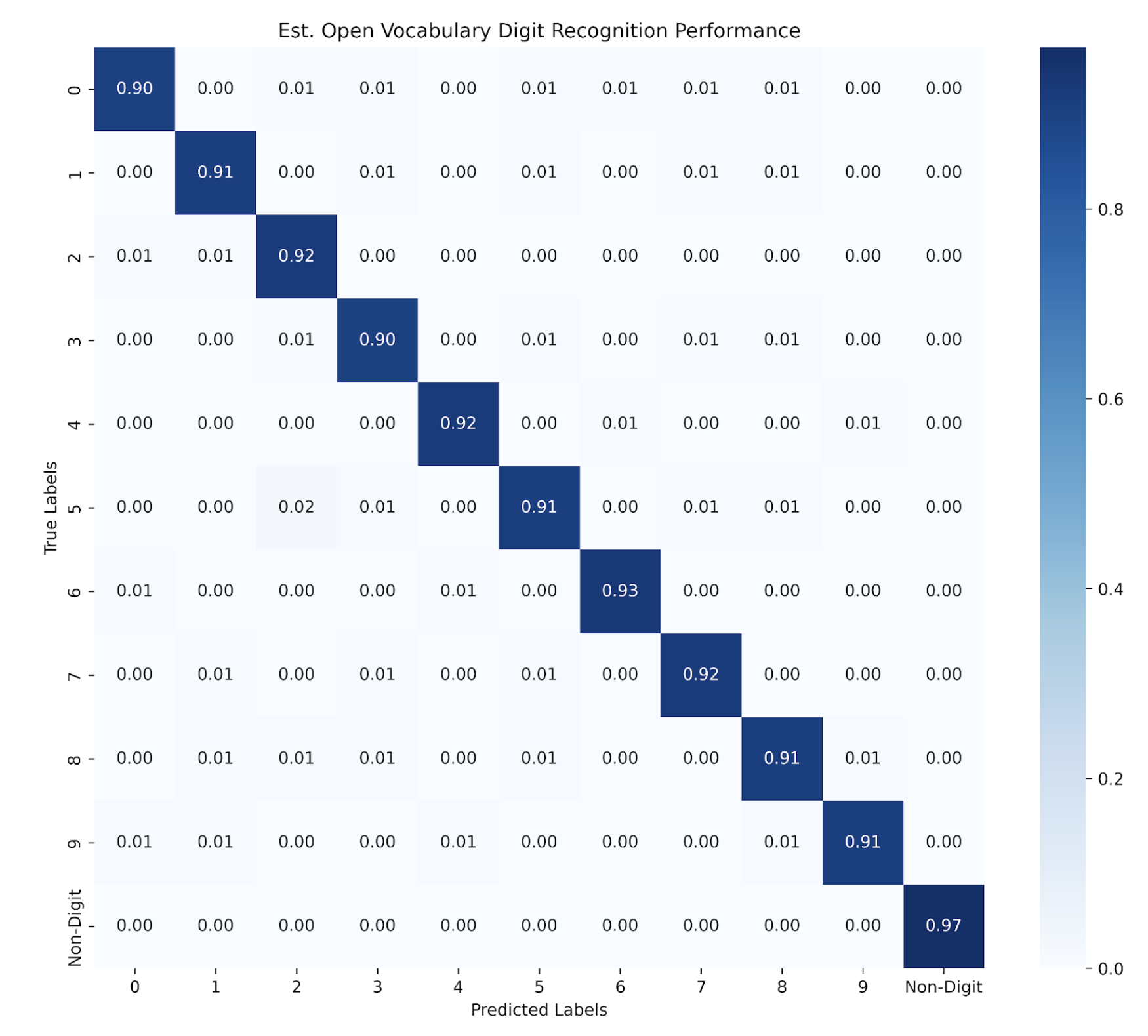}}
\end{figure}

The final 1D CNN classification model was evaluated on a lock box 100-sample test set comprising 20\% of the initial classification data. The lightweight, 5.2K-parameter classifier accepts the GCN-reduced \( \mathbb{R}^{256} \) output as input and applies two convolutional layers with max pooling and ReLU activation. The 1D CNN was optimized via Adam with a learning rate of 0.01 and trained over 70 epochs with a batch size of 80. The lightweight structure is intentional, as the model is meant to merely output a final classification with a robust, regularizable fit—the ``hard work" of classification is all handled upstream by the geometric machine learning pipeline. Indeed, the 1D CNN was assigned a fairly trivial classification task, as evident in \figureref{fig:image3}, which depicts a t-SNE projection of the manifold learning-generated GCN embeddings. In an attempt to ensure the integrity and reproducibility of our results, we did not alter any of the default scikit-learn settings for any of the t-SNE plots used in this study, thereby maintaining the standard configuration in order to provide unbiased representations of the data. 

In addition to the statistics reported in \sectionref{sec:Results}, more granular details regarding the final performance of the 1D CNN classifier—as a reflection of the upstream geometric ML processing—are described in \figureref{fig:image5}. By combining the final 1D-CNN with the claimed performance of the bidirectional RNN architecture from \citet{mahapatra} to determine final inter-digit granular classification, one could achieve an estimated \>90\% accuracy across all individual digits in an open-set learning environment. Note that this estimate, however, involves a balanced digit- versus non-digit dataset distribution, which may not generalize to real-world settings.

\section{AT-AT Details}\label{apd:third}

We provide further detail on the upstream autoencoder-targeted adversarial transformer (AT-AT) denoising model below. As previously mentioned, the autoencoder architecture \citep{choi,leite} follows a convolutional structure with batch normalization and dropout layers to stabilize training and prevent overfitting. Training is conducted via a correlation coefficient-driven objective function. The encoder starts with an input layer that accepts a 1D signal consisting of 512 frames, representing the EEG signal from a single channel. The encoder includes two convolutional layers: the first Conv1D layer contains 32 filters with a kernel size of 3, followed by batch normalization and ReLU activation. The second Conv1D layer has 64 filters with similar activation and normalization steps. Max pooling with a factor of 2 is applied after each convolution to reduce the temporal dimension. The central layer has 128 filters, maintaining the same structure of batch normalization and ReLU activation. On the decoder side, the architecture mirrors the encoder, with upsampling layers replacing the pooling layers. Two convolutional layers reconstruct the signal, and a final Conv1D layer with a single filter and sigmoid activation outputs the final reconstructed signal. The autoencoder is trained via Adam with a learning rate of $1 \times 10^{-4}$ and a batch size of 20 across ten epochs.

\begin{figure}[tb]
\floatconts
  {fig:image6}
  {\caption{Architectural diagram of the autoencoder-targeted adversarial transformer. The autoencoder and discriminator are in the lineage of \citet{choi}; an expanded treatment of AT-AT in isolation is given in \citet{choi2}.}}
  {\includegraphics[width=0.89\linewidth]{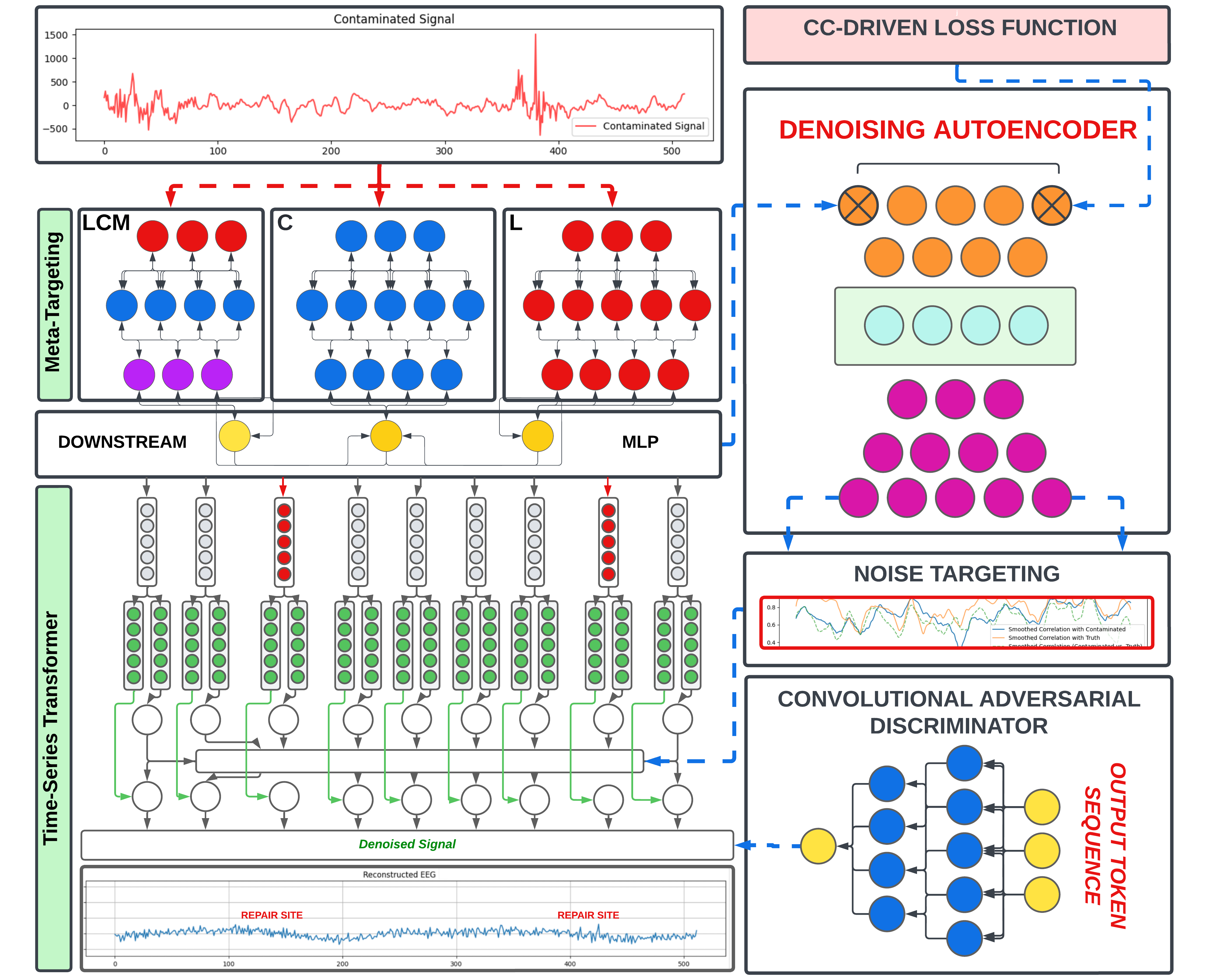}}
\end{figure}

\begin{figure}[h]
  \centering
  \includegraphics[width=0.95\columnwidth]{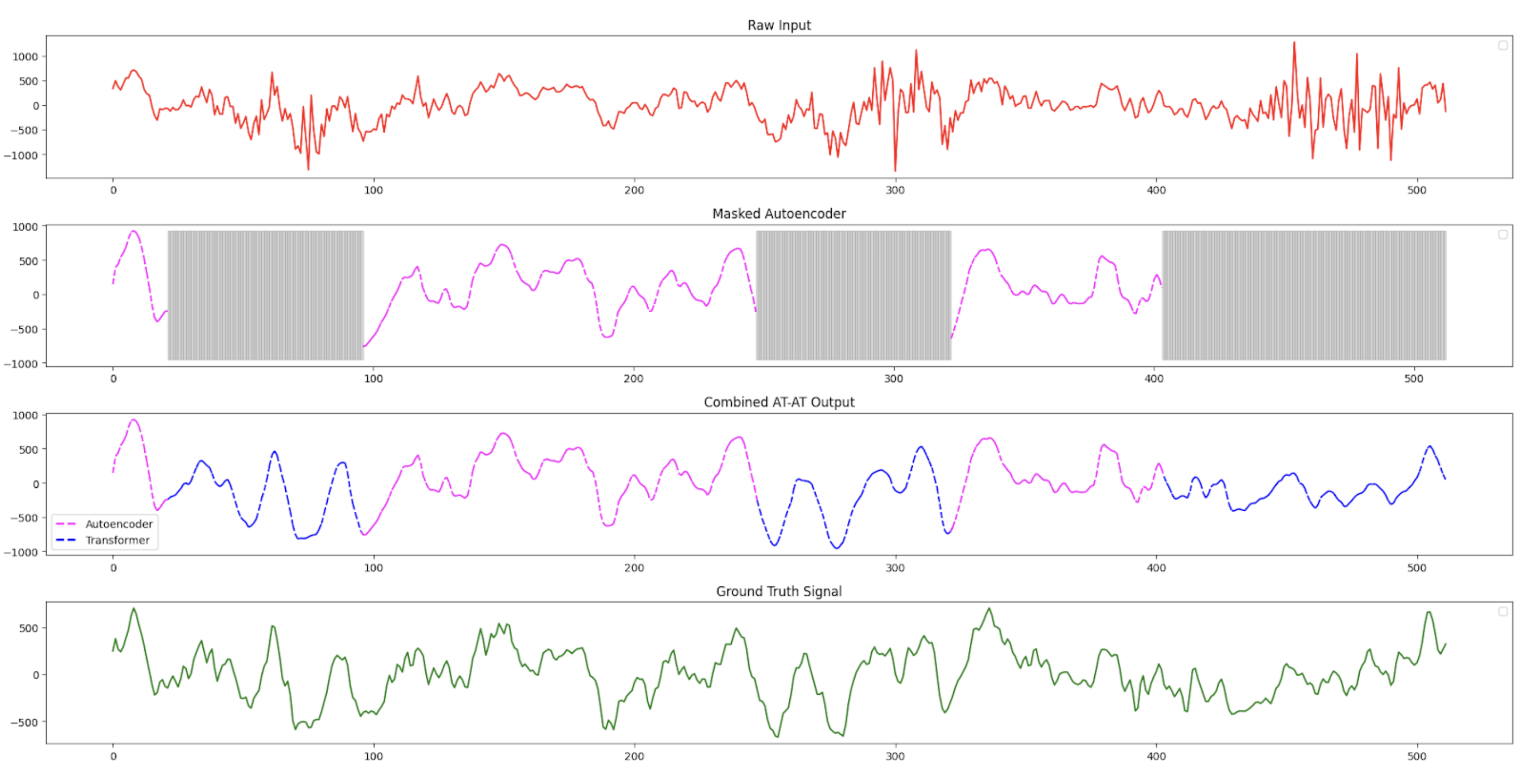}
  \caption{AT-AT model workflow; also given in \citet{choi2}. From top to bottom: (1) raw input signal, (2) initial autoencoder filtration pass with high-noise target site masking, (3) time-series transformer-based reconstruction of target sites, (4) ground truth signal.}
  \label{fig:image7}
\end{figure}

As mentioned in \sectionref{sec:Methods}, we trained an upstream pre-processing LSTM-CNN (LC) model to infer the appropriate SNR target level before deploying AT-AT, following previously demonstrated abstraction methods \citep{choi}. This model selects the suitable iteration of AT-AT based on the detected SNR level, enabling more tailored processing. Pre-processing accounted for 32.9\% of the 249-second training time. The LC model uses a hybrid architecture combining LSTM and CNN substructures to classify synthetic EEG and EMG data labeled with various SNR levels. The CNN pathway reshapes the EEG input into 2D blocks and processes them through Conv2D layers with ReLU, batch normalization, max-pooling, and dropout. The LSTM pathway captures temporal dependencies from the raw EEG data through two LSTM layers, which are flattened. The LSTM-CNN-MLP pathway involves two LSTM layers reshaped for convolutional processing via Conv2D followed by a dense MLP layer. These outputs are concatenated into a meta-classifier with two fully connected layers and a softmax output, which predicts the appropriate SNR level across the -7 to 2 dB range. The training process spans 100 epochs with a batch size of 100. Inferring relative SNR interference has been shown to be a relatively trivial task, with past accuracies demonstrated at 98\% \citep{soroush}; our upstream LC model was able to correctly infer SNR across all 100 test cases. (This pre-processing layer is used to toggle between downstream models and does not directly interact with the reconstruction process.)

The adversarial model uses the previously described generative adversarial network (GAN) architecture, combining a transformer-based generator and a convolutional discriminator to perform EEG denoising. The generator utilizes a transformer encoder with two encoder layers, multi-head attention (with four heads), and a feedforward network of size 128. Input features, which consist of two channels, are embedded into a 16-dimensional space before entering the transformer. The output from the transformer passes through a convolutional smoothing layer (Conv1D with a kernel size of 3); a fully connected layer then transforms the output into a 1D signal. The discriminator is a Conv1D-based model designed to distinguish between real and generated EEG signals. It contains two convolutional layers: the first Conv1D layer has 64 filters and LeakyReLU activation, followed by dropout, while the second layer has 128 filters and the same activation and dropout sequence. A fully connected layer reduces the representation to a scalar, which is processed by a sigmoid activation function to classify the signal as authentic or generated. The GAN training process alternates between generator and discriminator updates over five cycles in each iteration, with both models using binary cross-entropy loss for optimization. The Adam optimizer is used for both the generator and discriminator, with a learning rate of $1 \times 10^{-4}$. The adversarial model is trained over ten epochs with a batch size of 20.
\end{document}